\pdfoutput=1
\documentclass[11pt]{article}
\usepackage{float}

\usepackage[final]{acl}

\usepackage{times}
\usepackage{latexsym}

\usepackage[T1]{fontenc}
\usepackage[utf8]{inputenc}

\usepackage{microtype}

\usepackage{inconsolata}

\usepackage{graphicx}

\usepackage{listings}
\usepackage{xcolor}
\usepackage{hyperref}
\usepackage{enumitem}

\lstdefinestyle{json}{
    basicstyle=\ttfamily\footnotesize,
    numbers=left,
    numberstyle=\tiny,
    stepnumber=1,
    numbersep=5pt,
    backgroundcolor=\color{gray!10},
    showstringspaces=false,
    breaklines=true,
    frame=single
}

\title{Exploiting contextual information to improve stance detection in informal political discourse with LLMs\thanks{Dataset: \url{https://github.com/tonymullen/politics.com}. Code: \url{https://github.com/armanengin/contextual-stance-llms}.}}

\author{
 \textbf{Arman Engin Sucu\textsuperscript{1}},
 \textbf{Yixiang Zhou\textsuperscript{2}},
 \textbf{Mario A. Nascimento\textsuperscript{2}},
 \textbf{Tony Mullen\textsuperscript{1}}
\\ 
Khoury College of Computer Sciences\\
Northeastern University
\\
\textsuperscript{1}Seattle Campus, USA~~
\textsuperscript{2}Vancouver Campus, Canada
\\
  \small{
    \textbf{Correspondence:} \href{mailto:a.mullen@northeastern.edu}{a.mullen@northeastern.edu}
  }
}

\begin{document}
\maketitle
\begin{abstract}
This study investigates the use of Large Language Models (LLMs) for political stance detection in informal online discourse, where language is often sarcastic, ambiguous, and context-dependent. We explore whether providing contextual information, specifically user profile summaries derived from historical posts, can improve classification accuracy. Using a real-world political forum dataset, we generate structured profiles that summarize users' ideological leaning, recurring topics, and linguistic patterns. We evaluate seven state-of-the-art LLMs across baseline and context-enriched setups through a comprehensive cross-model evaluation. Our findings show that contextual prompts significantly boost accuracy, with improvements ranging from +17.5\% to +38.5\%, achieving up to 74\% accuracy that surpasses previous approaches. We also analyze how profile size and post selection strategies affect performance, showing that strategically chosen political content yields better results than larger, randomly selected contexts. These findings underscore the value of incorporating user-level context to enhance LLM performance in nuanced political classification tasks.
\end{abstract}

\section{Introduction}

Political stance detection is an increasingly relevant part of analyzing the flow of ideas in online environments where discourse is informal and implicitly expressed. Understanding a text or individual's ideological standpoint can be helpful for applications such as content moderation, public opinion tracking, and misinformation detection. Approaches to political stance detection using traditional natural language processing (NLP) and machine learning methods have been closely related to approaches to sentiment analysis. 

However, political language is often nuanced and tends to be comparable to relatively difficult sentiment analysis domains. Posts with political stance on social networks are often ambiguous, sarcastic, or context-dependent. For example, consider the statement: \textit{"Great, another tax cut for the rich—just what we needed!"}. Without additional context, this could either express support or sarcasm. Political intent is often embedded in subtext or prior engagement, which traditional models fail to capture \cite{mullenMalouf2006, malouf2008taking, samih-darwish-2021-topical}.

While earlier methods such as lexicon-based classifiers or keyword matching approaches perform poorly on such nuanced input, recent advancements in LLMs such as GPT-4~\cite{openai-gpt4o}, LLaMA~\cite{llama3}, and DeepSeek~\cite{deepseekai2025deepseekr1} offer promise in handling complex language understanding \cite{cao2024metadata, kim2024robust}.

The emergence of LLMs has fundamentally transformed approaches to sentiment analysis and stance detection. Traditional methods based on lexicons, feature engineering, and specialized classifiers have been largely supplanted by these general-purpose models that can capture subtle linguistic nuances, contextual cues, and implicit sentiment without task-specific architectures \cite{cruickshank2024prompting, allaway2023zeroshot}. However, despite this paradigm shift, the core challenge of contextual understanding remains \cite{bhattacharya2024userstance}.

Nonetheless, even state-of-the-art LLMs struggle with implicit political signals, ideological ambiguity, and sarcastic cues. Our project investigates whether political stance can be reliably classified by augmenting LLM predictions with contextual cues, building on previous research that demonstrated the value of contextual information in political classification tasks \cite{mullenMalouf2006, malouf2008taking, doddapaneni-etal-2024-user}.

In this study, we introduce a contextual enrichment framework that supplements LLM input with user profile summaries derived from historical forum posts. These profiles include inferred political leaning, recurring discussion topics, and linguistic patterns \cite{wu2024understanding, ye-etal-2021-beyond}. By providing this additional context, we aim to improve stance classification accuracy—especially for posts that are short, ambiguous, or stylistically neutral.

We evaluate this approach on a real-world political forum dataset, comparing baseline classification against context-enhanced setups through a comprehensive cross-model evaluation of seven state-of-the-art LLMs. Our results show that incorporating profile-level context significantly improves model performance, with absolute accuracy gains ranging from +24.5\% to +38.5\%. We further investigate how profile size and post selection strategies affect performance, revealing that strategically selected political content contributes more than sheer volume \cite{cao2024metadata, welch-etal-2022-leveraging}.

This work highlights the importance of integrating user-level context into prompt design for political NLP tasks and offers a scalable method for enhancing classification reliability in informal discourse settings.

\section{Related Work}

Political stance detection spans multiple research traditions, from early sentiment analysis to recent LLM-based approaches. We review work in three key areas: (1) political stance classification techniques, (2) contextual enrichment methods, and (3) personalization for language models.

\subsection{Political Stance Classification}
Political sentiment analysis has long informed efforts to identify ideological positions in text. Early work focused on classifying opinion polarity in political tweets or news, often using lexicons or shallow models \cite{Mohammad2017StanceSentiment, Caetano2018Homophily}. Studies also highlighted the role of affect in political discourse and the asymmetry of negative sentiment spread \cite{Antypas2023Negativity, sen-etal-2020-reliability}. More recent research developed domain-specific and multilingual models to better capture political meaning in social media content \cite{Aquino2025GraphAware, kawintiranon-singh-2022-polibertweet}.

Building on this foundation, political stance detection has progressed from rule-based and lexicon-driven methods to neural and prompt-based approaches. Early studies explored user-level classification in online forums using discourse features \cite{mullenMalouf2006, malouf2008taking, samih-darwish-2021-topical, zhou2024silent}, highlighting challenges posed by implicit and informal political language. While these approaches laid important groundwork for modeling user-level political stance, they lacked the contextual understanding capabilities that our approach leverages.

\subsection{Contextual LLM Approaches}
Recent LLMs enable zero- and few-shot stance classification without task-specific models. Prompting strategies with metadata or topic cues improve accuracy \cite{cao2024metadata, cruickshank2024prompting, kim2024robust, allaway2023zeroshot}. User-level modeling further boosts performance by leveraging behavioral or linguistic summaries \cite{bhattacharya2024userstance, doddapaneni-etal-2024-user, welch-etal-2022-leveraging, wu2024understanding, ye-etal-2021-beyond}. Evaluations on social media platforms like Twitter/X demonstrate model potential and limitations \cite{Gambini2024StanceLLM}, while frameworks like DEEM dynamically adapt to user history \cite{wang-etal-2024-deem}. Our work extends these approaches by systematically exploring how different types of user-level context affect classification accuracy across diverse LLM architectures.

\subsection{Personalization and Reasoning in LLMs}
Personalization in LLMs has advanced through techniques such as persona-aware attention, guided profile generation, retrieval-augmented prompting, and adaptive calibration. These methods have shown strong performance across dialogue, writing assistance, and recommendation tasks~\cite{huang2023persona, zhang2024guided, salemi-etal-2024-lamp, tan-etal-2024-personalized, mysore2024pearl}. Recent work also highlights the importance of preference alignment, with studies evaluating how well LLMs follow user-specific instructions in downstream tasks~\cite{Zhao2025PrefEval}.

Complementary to these personalization efforts, recent research has explored reasoning-aware prompting strategies—such as Chain-of-Thought~\cite{wei2022cot, kojima2022large}, ReAct~\cite{yao2023react}, AutoPrompt~\cite{shin-etal-2020-autoprompt}, and prefix-tuning~\cite{li2021prefix}—which aim to improve model understanding of implicit, ambiguous, or sarcastic cues. While our approach does not employ these methods, they represent promising future directions. Techniques like Chain of Preference Optimization~\cite{Zhang2024CPO}, which integrate user preferences into multi-step reasoning, may further enhance stance detection when combined with contextual enrichment.

Our work focuses specifically on user-level contextual prompting. By enriching model input with structured user profiles, we show consistent improvements in stance classification across seven state-of-the-art LLMs. These findings highlight the value of user-informed prompting in capturing nuanced signals in political discourse, and they may complement reasoning-based approaches in future hybrid systems.

\section{Dataset and Preprocessing}

Our study utilizes a political discourse dataset originally compiled by \citet{mullenMalouf2006}, consisting of approximately 77,854 posts downloaded from discussions on politics.com. The dataset is organized into topic threads, chronologically ordered, and identified according to author and author's stated political affiliation.

\subsection{Data Source and Characteristics}

The dataset contains contributions from 408 unique users engaged in various political discussions. User posting activity follows an inverse power-law distribution typical of online communities, with 77 posters (19\%) contributing only a single post. The most active user contributed 6,885 posts, followed by the second most active with 3,801 posts.

A key feature of this dataset is that users self-declared their political affiliations, providing ground truth labels for our classification task.

Figure \ref{fig:distribution} shows the distribution of political affiliations in the dataset, which is relatively balanced between major ideological groups.

\begin{figure}[h]
\centering
\begin{tabular}{cll}
\hline
                & Republican   & 53 \\
RIGHT 34\%      & Conservative & 30 \\
                & R-fringe     & 5  \\
\hline
                & Democrat     & 62 \\
LEFT 37\%       & Liberal      & 28 \\
                & L-fringe     & 6  \\
\hline
                & Centrist     & 7  \\
                & Independent  & 33 \\
OTHER 28\%      & Libertarian  & 22 \\
                & Green        & 11 \\
\hline
                & Unknown      & 151 \\
\hline
\end{tabular}
\caption{Distribution of posts in the data by general class and by a slightly modified version of the writers' own self-descriptions.}
\label{fig:distribution}
\end{figure}

\subsection{Data Preprocessing}
For our experiments, we processed this dataset in several key ways:
\begin{enumerate}
    \item We mapped the original fine-grained political affiliations into three broad categories: LEFT (Democrat, Liberal, Left-fringe), RIGHT (Republican, Conservative, Right-fringe), and UNKNOWN (all other labels including Centrist, Independent, Libertarian, and Green).
    
    \item We focused only on users with clear LEFT or RIGHT labels, filtering out posts from users with UNKNOWN political affiliation. This resulted in a filtered dataset of 56,035 posts from 257 users with declared political leanings.
    
    \item For each user with a known political affiliation, we split their posts into two sets: 70\% for profile generation (used to create user context) and 30\% for testing classification performance (reserved for evaluation). We used a fixed random seed (42) for this split to ensure reproducibility across experiments and enable direct comparison of results.
    
    \item We maintained post structure and metadata throughout preprocessing by preserving quote markers to differentiate between original content and quoted text, keeping forum-specific formatting to maintain conversational context, and retaining chronological ordering within each user's posts.
\end{enumerate}
This approach allowed us to maintain the informal, conversational nature of the discourse while creating a structured dataset suitable for both baseline and context-enriched classification experiments. To ensure experimental rigor, we used the same test set for all experiments, allowing direct comparison between baseline and context-enhanced approaches.

\section{Methodology and Experimental Design}
\label{sec:methodology}

Our approach centers on how contextual information about users' past behaviors can enhance LLMs' ability to classify political stance in informal discourse. We conducted three distinct experiments to thoroughly investigate the effectiveness of contextual enrichment.

\subsection{Experimental Framework Overview}

\subsubsection{Implementation Approach}
We define this as a binary stance classification task. Each input consists of a single forum post authored by a user. In the baseline setup, the post is provided to the model in isolation. In the context-enriched setup, the same post is preceded by a structured user profile summarizing the author’s historical political behavior. The model is prompted to return a JSON object containing a predicted stance label—LEFT or RIGHT—and an accompanying explanation. The ground truth label is derived from the user’s self-declared political affiliation in the dataset.

All experiments shared a common implementation approach to ensure consistent results. We accessed the LLMs through a unified API interface, providing standardized access across different model architectures. To maintain consistency, we applied identical parameters across all experiments: temperature set to 0.1 to minimize stochastic variation, standardized JSON output format for automated evaluation, and identical prompt structures except for the addition of context. Throughout our experiments, we evaluated two classification pipelines: a baseline where models classify posts without any user context, and a context-enriched approach where the same posts are classified with user profiles prepended in the prompt.

\subsubsection{Experimental Progression}
We implemented three sequential experiments, with each building on findings from the previous:
\begin{enumerate}
    \item \textbf{Contextual Enrichment Impact}: Evaluating the maximum potential benefit of user profiles for classification accuracy
    \item \textbf{Context Optimization Framework}: Determining optimal post selection strategies and volume for profile generation
    \item \textbf{Cross-Model Performance Analysis}: Assessing different LLMs' capabilities in both profile generation and classification roles
\end{enumerate}

\subsection{User Profile Structure}
Across all experiments, we used a consistent structured format for user profiles. Each profile contained the inferred political stance (left, right, or unknown) based on consistent ideological signals, the model's self-assessed confidence in its stance assignment (high, medium, or low), 3–5 specific linguistic or topical indicators supporting the assigned leaning, a list of common subjects the user discusses, a qualitative summary of the user's tone, a description of whom the user supports or criticizes, and optional free-text insights. These fields were generated using a structured prompt (see Appendix~\ref{appendix:User Profile Summarization Prompt}), emphasizing objectivity, pattern recognition, and valid JSON formatting.

\subsection{Experiment 1: Contextual Enrichment Impact}
Our first experiment aimed to establish whether user profiles could improve classification performance and to measure the maximum potential benefit. We used Gemini 2.0 Flash~\cite{gemini-flash} (with its 1M token context window) to generate comprehensive user profiles from all available posts in the profile-building set. Unlike later experiments, we did not selectively sample posts but instead used all available posts per user to generate the most comprehensive profiles possible. We evaluated on a set of 200 reserved test posts, ensuring a balanced representation of different political orientations. This experiment established the ceiling performance for our contextual enrichment approach.

\subsection{Experiment 2: Context Optimization Framework}
After establishing the effectiveness of contextual enrichment, we investigated how to optimize the context generation process. We implemented and evaluated five distinct post selection strategies:
The \textbf{PoliticalSignalSelection} strategy prioritizes posts with strong political content by using a weighted lexicon of political terms in three categories: general political terms (e.g., 'politics', 'government', 'vote') with weight 1, party-specific terms (e.g., 'democrat', 'republican', 'liberal') with weight 2, and hot-button issues (e.g., 'abortion', 'gun', 'immigration') with weight 3. It calculates a political signal score for each post based on term frequency, boosts scores for posts in political subforums (+5 points), adds small random noise (0–1) to break ties, and selects 60\% highest-scoring posts and 40\% diverse-topic posts (see Appendix~\ref{appendix:selection-strategies} for full implementation details).

Prior studies have shown that content-based filtering—specifically targeting politically salient posts—significantly improves stance detection accuracy. \citet{aldayel2019stance} and \citet{preotiuc-pietro-etal-2017-twitter} demonstrate that features derived from political lexicons and issue-related keywords are more informative than post recency or length. \citet{rahimzadeh2025userprofiling} further showed that filtering timelines to remove off-topic content enhanced user profiling with LLMs in large-scale settings. These findings support our PoliticalSignalSelection strategy: by scoring and selecting posts with high ideological signal, we retain the most diagnostic content for modeling user stance.

We also tested \textbf{RandomSelection} (randomly samples posts without consideration for content), \textbf{ControversialTopicSelection} (prioritizes posts containing terms from contentious political topics using a library of 150+ controversial keywords), \textbf{RecentPostSelection} (selects the most recent posts from a user's history), and \textbf{LongFormSelection} (prioritizes longer posts based on word count).

We evaluated eight different post count settings to understand the relationship between context volume and classification performance, ranging from minimal context (1, 2, 3 posts), medium context (5, 10 posts), and extensive context (20, 30 posts), to maximum context (50 posts). We tested each combination of post count and selection strategy, resulting in 40 distinct experimental conditions (8 post counts × 5 selection strategies). Each condition was tested on up to 50 users with 5 test posts per user (max 250 classification instances per condition), for a total of approximately 10,000 classification instances across all conditions.

Through this experiment, we determined that \textbf{PoliticalSignalSelection} with 10-20 posts yielded near-optimal results, with diminishing returns beyond this threshold.

\subsection{Experiment 3: Cross-Model Performance Analysis}
Our final experiment investigated how different LLMs perform in both profile generation and classification roles, using the optimized parameters from Experiment 2. We tested seven state-of-the-art LLMs representing diverse architectures: Claude 3.7 Sonnet~\cite{claude-sonnet}, Grok-2-1212B~\cite{grok2}, GPT-4o Mini~\cite{openai-gpt4o}, Mistral Small-24B~\cite{mistral-small}, Meta-LLaMA 3.1-70B~\cite{llama3}, Qwen~\cite{qwen2.5}, and Gemini 2.0 Flash~\cite{gemini-flash}.

Based on findings from Experiment 2, we standardized parameters across models, using only the PoliticalSignalSelection strategy, 50 posts per user profile, and the same test dataset of 200 posts per model. We implemented a 7×7 experimental design where each model generated user profiles for the same set of users, each model was then used to classify posts using profiles created by every model, and all 49 model combinations were evaluated using the same test dataset. This comprehensive evaluation revealed which models excel at generating informative profiles and which are most effective at leveraging contextual information for classification.

\subsection{Evaluation Approach}
To assess the impact of contextual enrichment across our experiments, we focused on several key comparative metrics. We measured absolute improvement as the percentage point difference between context-enriched and baseline accuracy, directly quantifying the benefit of providing user profiles. We analyzed the relative impact across models by examining how improvement correlates with baseline performance, revealing whether weaker models benefit more from contextual information. We studied context efficiency as performance relative to context volume, helping identify the optimal balance between context size and computational requirements. Finally, we analyzed cross-model complementarity, determining which model combinations (profile generator + classifier) yield the best performance and reveal potential complementary strengths.

\section{Results and Analysis}
\subsection{Contextual Enrichment}
To address the challenge of stance ambiguity in informal political discourse, we explored whether providing contextual information about users could improve classification accuracy. This approach extends the work of \citet{malouf2008taking}, who achieved 68.48\% accuracy using graph-based social context (who quotes whom) combined with Naive Bayes classification. Our research investigates whether user profile summaries can provide similar contextual benefits when applied to modern LLMs. We tested seven different LLMs on the same dataset with and without user profile summaries.
\subsubsection{Impact of User Profiles on Classification Accuracy}
\begin{figure*}[h]
    \centering
    \includegraphics[width=\textwidth]{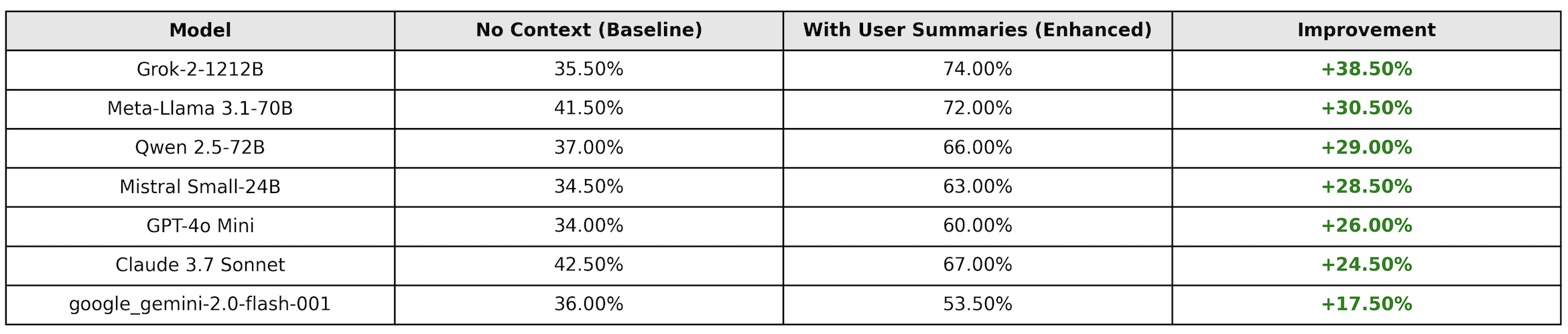}
    \caption{Classification accuracy comparison with and without user profile summaries.}
    \label{fig:context_improvement}
\end{figure*}
Figure~\ref{fig:context_improvement} demonstrates that adding user profile summaries substantially enhances stance classification across all models tested. This contextual enrichment approach produced significant improvements that ranged from +17.50\% to +38.50\% in absolute precision.

The most striking improvement was observed with Grok-2-1212B, which saw a +38.50\% increase (from 35.50\% to 74.00\%). Despite having a relatively low baseline performance, this model exhibited the greatest benefit from contextual information. The Meta-Llama 3.1-70B model, while starting from a higher baseline (41.50\%), still achieved a substantial +30.50\% improvement when provided with user summaries.

Even the model with the highest baseline accuracy, Claude 3.7 Sonnet (42.50\%), gained a significant +24.50\% improvement with context enhancement. Google's Gemini 2.0 Flash showed the most modest improvement at +17.50\%, which aligns with a broader pattern we explore in Section~\ref{sec:optimal_model_combinations}, where we discover that models often perform sub-optimally when classifying using their own generated profiles compared to profiles generated by other models. Despite Gemini being a competent classifier overall, this particular limitation affected its performance in this experiment.
To explore the maximum potential of our approach, we used all available posts except the 200 reserved for testing to generate the most comprehensive user profiles possible, which led to our peak accuracy of 74.00\% with Grok-2-1212B.

Notably, our highest accuracy result (74.00\% with Grok-2-1212B) surpassed the best result from \citet{malouf2008taking} (68.48\%), despite our approach using a different form of contextual information. This indicates that LLMs with user profiles can effectively leverage context in ways comparable to or better than traditional methods using explicit social network information.
\subsubsection{Context Size and Selection Strategy}

\begin{figure}[h]
    \centering
    \includegraphics[width=\linewidth]{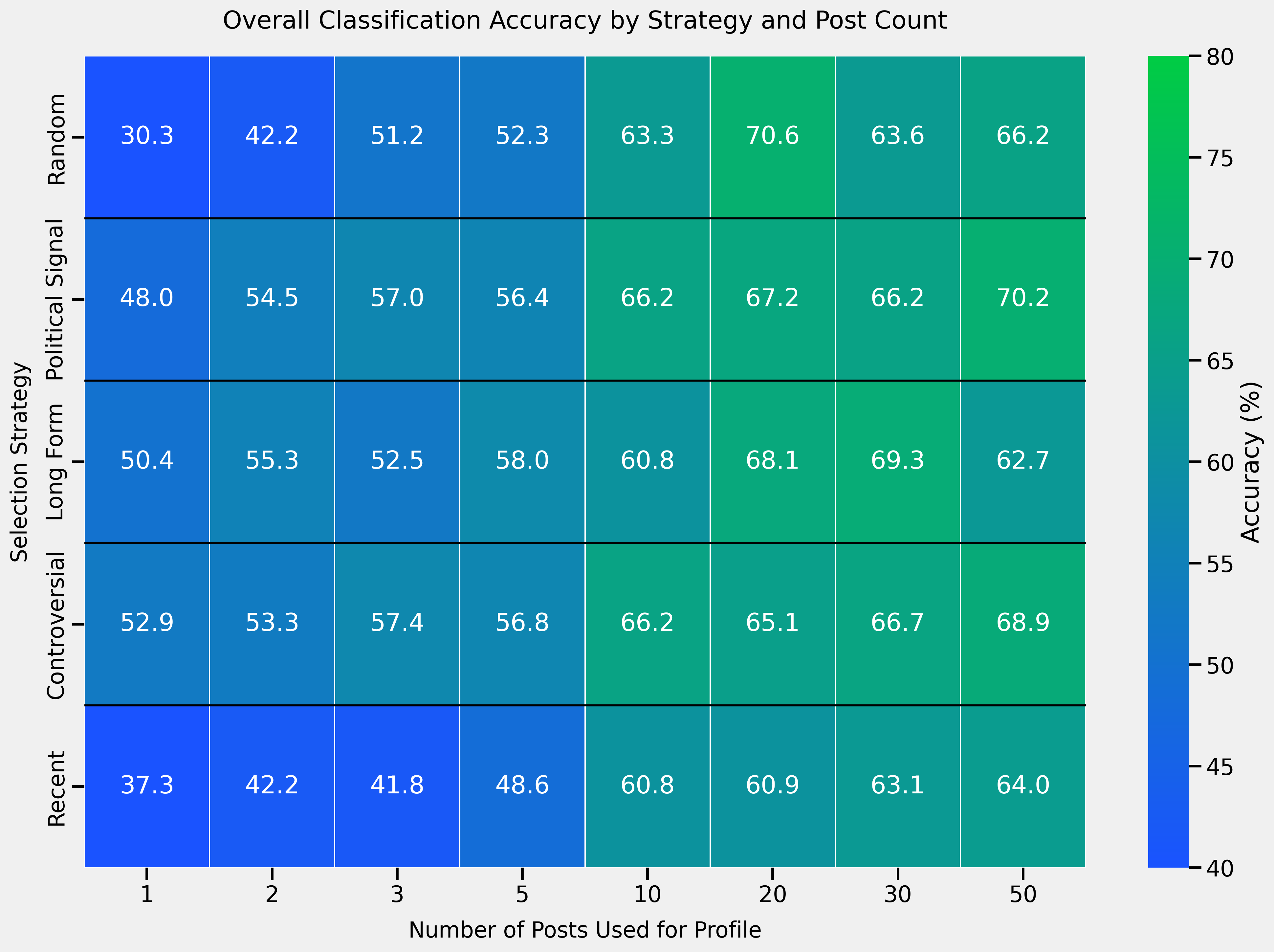}
    \caption{Accuracy by post selection strategy and number of posts used for user profiles.}
    \label{fig:context_strategy}
\end{figure}

Our earlier experiments (Figure~\ref{fig:context_strategy}) reveal that both the quantity and selection strategy of posts used to create user profiles significantly impact classification performance. When comparing different post selection strategies, we found that sampling based on \textbf{political signal strength} generally outperformed other approaches, reaching 70.2\% accuracy when using 50 posts per user.

However, the relationship between post count and accuracy is non-linear. We observed diminishing returns after 10-20 posts, with most strategies showing only modest gains beyond this threshold. For instance, the political signal strategy achieved 66.2\% accuracy with just 10 posts, which increased only marginally to 70.2\% with 50 posts.

Interestingly, the random selection strategy showed the most substantial gains when scaling from 10 posts (63.3\%) to 20 posts (70.6\%), suggesting that volume can partially compensate for less sophisticated selection methods. However, its performance declined with higher post counts, potentially due to the inclusion of irrelevant content that dilutes relevant signals.

These findings indicate that while providing more context generally improves performance, strategic selection of highly relevant posts yields better results than simply increasing context volume. This has important implications for real-world applications, where processing efficiency must be balanced against classification accuracy.

\subsubsection{Cross-Model Applicability}

An important question is whether contextual enrichment benefits all models equally or if certain architectures are better suited to leveraging user profile information. Our experiments show that while all models improved significantly, the relative gains were inversely proportional to baseline performance. Models with weaker baseline performance (Grok, Qwen, Mistral, GPT-4o Mini) saw the largest relative improvements, suggesting that contextual information may have a normalizing effect—bringing underperforming models closer to the capabilities of stronger ones.

This pattern indicates that contextual enrichment is particularly valuable for deployment scenarios where computational constraints necessitate using smaller or less capable models. By providing well-curated user profiles, even models with limited parameters can achieve competitive stance classification performance.

\subsection{Cross-Model Performance Analysis}

\begin{figure}[h]
    \centering
    \includegraphics[width=\linewidth]{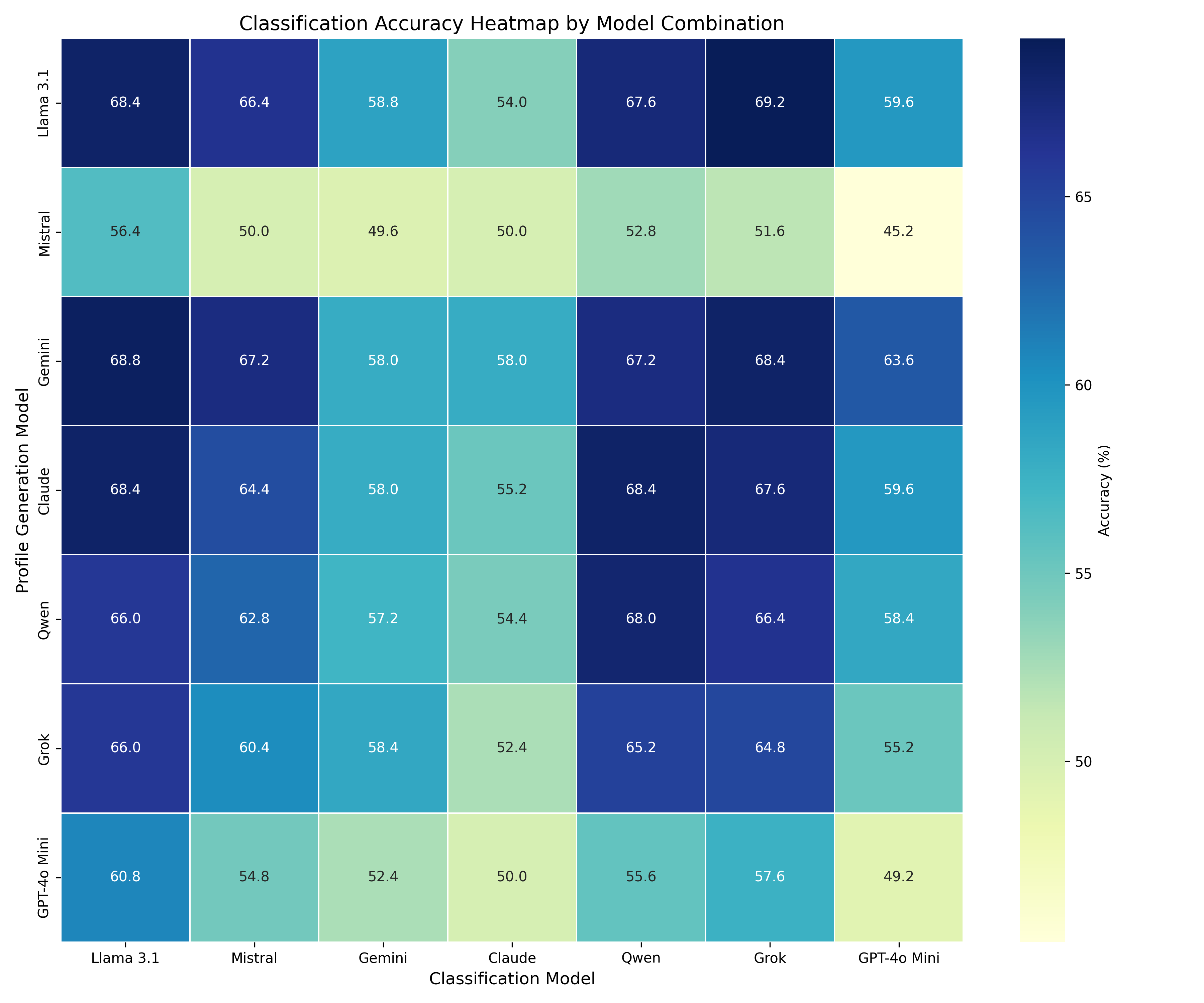}
    \caption{Classification accuracy heatmap by model combination. Profile generation models are shown on the y-axis, while classification models are on the x-axis.}
    \label{fig:cross_model_heatmap}
\end{figure}

To understand the relative strengths of different LLMs in the context-enriched classification pipeline, we conducted a comprehensive cross-model evaluation. As shown in Figure~\ref{fig:cross_model_heatmap}, we tested all combinations of profile generation and classification models, revealing several important patterns:

\subsubsection{Profile Generation Capabilities}
The vertical dimension of the heatmap reveals which models excel at generating informative user profiles. Our analysis shows that Llama 3.1, Gemini, Claude, Qwen, and Grok consistently produce high-quality profiles, enabling classification accuracies above 60\% when used with strong classification models. In contrast, Mistral Small and GPT-4o Mini demonstrate weaker profile generation capabilities, with their profiles resulting in generally lower classification accuracy across all classification models. Notably, Llama 3.1 profiles yield the best overall performance, with an average accuracy of 63.4\% across all classification models, suggesting superior capability in distilling relevant political patterns from user post history.

\subsubsection{Classification Strengths}
The horizontal dimension of the heatmap reveals which models most effectively utilize profile information for classification. Llama 3.1 and Grok stand out as the strongest classification models, achieving high accuracy regardless of which model generated the profiles. Claude and Gemini demonstrate middling performance as classifiers, while still benefiting significantly from high-quality profiles. In contrast, GPT-4o Mini consistently performs weakest as a classifier across most profile sources, suggesting potential limitations in its ability to interpret and apply contextual information.

\subsubsection{Optimal Model Combinations}\label{sec:optimal_model_combinations}
The most effective combinations revealed by our experiments were Gemini + Llama 3.1 (68.8\% accuracy), Llama 3.1 + Grok (69.2\% accuracy), and Claude + Qwen (68.4\% accuracy). Interestingly, we found that most models perform better when using profiles generated by a different model rather than their own profiles (the diagonal is not consistently highest). This suggests complementary strengths between different models in the context-enriched classification pipeline. For example, while Llama 3.1 is strong in both roles, it achieves its peak performance (69.2\%) when classifying posts using Grok-generated profiles rather than its own.

This finding has important practical implications, suggesting that hybrid approaches combining different models for profile generation and classification may yield better results than using a single model for the entire pipeline.

\subsection{Synthesis of Findings}

Our experiments reveal three key insights that advance our understanding of political stance classification in informal discourse:

\begin{enumerate}
    \item \textbf{Contextual enrichment significantly improves performance} across all models tested, with absolute accuracy gains of +17.50\% to +38.50\%. This confirms and extends \citet{malouf2008taking}'s finding that contextual information is crucial for this task.
    
    \item \textbf{Strategic post selection is more important than quantity} when building user profiles. The political signal selection strategy with just 10-20 posts can achieve nearly optimal performance, offering an efficient approach for real-world applications.
    
    \item \textbf{Different models exhibit complementary strengths} in the profile generation/classification pipeline, with the best results achieved by combining models that excel in each respective role.
\end{enumerate}

These findings demonstrate that modern LLMs can effectively leverage user context for political stance classification, achieving results comparable to or better than traditional methods using explicit social network information. Furthermore, our work reveals that careful optimization of contextual information and model selection can substantially enhance performance on this challenging task.

\section{Conclusion}

In this paper, we investigated how LLMs can be leveraged to accurately classify political stances in informal discourse by incorporating user-level contextual information. Our research demonstrates that providing summarized user profiles based on historical posts significantly enhances classification accuracy across all tested models, with improvements ranging from +17.50\% to +38.50\%.

We found that strategic selection of posts with strong political signals yields better results than simply maximizing context volume, with diminishing returns observed beyond 10-20 posts per user. This suggests efficient approaches for real-world applications where processing constraints may limit context size. Our cross-model evaluation further revealed that different LLMs exhibit complementary strengths in the context-enriched classification pipeline, with some models excelling at profile generation while others perform better at classification.

Our best result—74.00\% accuracy with Grok-2-1212B using comprehensive user profiles—surpassed previous approaches that relied on social network information. This demonstrates that modern LLMs with appropriate contextual information can effectively address the challenge of political stance detection in informal, ambiguous discourse settings.

\section*{Limitations}
While our research demonstrates significant improvements in political stance classification through contextual enrichment, several limitations should be acknowledged:
(1) Our dataset from politics.com represents a specific time period and cultural context that predates current political divisions, potentially limiting direct applicability to contemporary discourse across different platforms and demographics;
(2) Our LEFT/RIGHT classification framework simplifies the spectrum of political ideologies, necessary for experimental clarity but not fully reflecting the complexity of real-world political stances;
(3) Practical constraints limited our testing of all possible combinations of model parameters, profile sizes, and prompt formulations. Future work could explore more nuanced political categorization beyond binary classification, test generalizability across diverse political discourse platforms, and investigate optimal context generation strategies for specific model architectures, potentially yielding even more accurate stance detection systems for real-world applications;
(4) While our method focuses on user-level contextual enrichment, we did not explore reasoning-aware prompting strategies such as Chain-of-Thought~\cite{wei2022cot}, ReAct~\cite{yao2023react}, or prefix-tuning~\cite{li2021prefix}. These techniques may help models better interpret sarcastic or implicit cues in political discourse, and their integration with user-informed prompting represents a promising direction for future research.

\section*{Ethical Considerations}
Our research on political stance classification raises several ethical considerations: (1) Dual-Use Potential: While intended to improve understanding of political discourse, these technologies could potentially be used for political profiling or surveillance, highlighting the importance of applications focused on enhancing communication rather than targeting individuals; (2) Algorithmic Bias: Stance classification systems may perpetuate biases present in training data or models, necessitating monitoring for systematic errors affecting specific political groups; (3) Transparency and Consent: Applications should clearly disclose how user data is processed and political stances are inferred, with appropriate opt-out mechanisms for users whose historical data is analyzed. We recommend that implementations be accompanied by oversight mechanisms and ethical guidelines that respect political diversity and user privacy, particularly in environments where political expression may carry social or professional consequences.

\bibliography{custom}

\appendix

\section{User Context and Profile Summarization} \label{appendix:user-context}

\subsection{User Profile Summarization Prompt} \label{appendix:User Profile Summarization Prompt}

\begin{quote}
Analyze the following set of forum posts by the user and create a concise political profile summary. For this task:
\begin{enumerate}
\item Identify any consistent political indicators in their posts (criticism of specific politicians/parties, stance on issues, etc.)
\item Note recurring topics this user discusses
\item Observe distinctive language patterns (formal/informal, emotional/detached, specific phrases)
\item Identify who/what they consistently criticize or support
\item Determine if there's sufficient evidence to classify them as LEFT, RIGHT, or UNKNOWN
\end{enumerate}

Format your response as a JSON object with these fields:
\begin{lstlisting}[style=json]
{
  "username": "the username",
  "political_leaning": "left/right/unknown",
  "confidence": "high/medium/low",
  "key_indicators": ["3-5 specific examples from posts that indicate political leaning"],
  "recurring_topics": ["list frequent topics"],
  "language_style": "brief description of their communication style",
  "sentiment_patterns": "who/what they criticize or support", 
  "context_notes": "any additional relevant information"
}
\end{lstlisting}

\textbf{IMPORTANT:}
\begin{itemize}
\item Focus on clear patterns rather than isolated statements
\item Maintain objectivity and avoid overinterpreting ambiguous content
\item If there isn't sufficient evidence to determine orientation, mark as ``unknown''
\item Ensure your response is a valid JSON object
\end{itemize}
\end{quote}

\subsection{Classification with Profile Summary Prompt} \label{appendix:Classification with Profile Summary Prompt}

\begin{quote}
Analyze the following discussion group post and classify the author's political orientation.

\textbf{IMPORTANT CONTEXT ABOUT THIS USER:}\\
\{profile\_summary\}

Take the above user profile into account when analyzing this post. The profile reflects patterns from the user's previous posts, which may provide context for this specific post.

Provide your response in this exact JSON format:

\begin{lstlisting}[style=json]
{
  "orientation": "LEFT|RIGHT|UNKNOWN",
  "explanation": "A detailed explanation of why you chose this classification based on the content"
}
\end{lstlisting}
\end{quote}

\section{Post Selection Strategy Implementation Details} \label{appendix:selection-strategies}

In this section, we provide the detailed implementation of our post selection strategies, particularly the \textbf{PoliticalSignalSelection} algorithm that performed best in our experiments.

\subsection{PoliticalSignalSelection Algorithm}

The \textbf{PoliticalSignalSelection} strategy uses a weighted lexicon approach to identify posts with strong political content. The algorithm works as follows:

\begin{enumerate}
\item \textbf{Term Weighting:} Political terms are categorized and weighted based on their signal strength:
   \begin{itemize}
   \item \textit{General political terms} (weight 1): 'politics', 'political', 'government', 'policy', 'policies', 'election', 'vote', 'voting', 'democracy', 'democratic'
   \item \textit{Party-specific terms} (weight 2): 'democrat', 'democratic party', 'liberal', 'progressive', 'socialism', 'left', 'left-wing', 'republican', 'gop', 'conservative', 'right', 'right-wing', 'trump', 'biden', 'obama', 'maga', 'tea party'
   \item \textit{Hot-button issues} (weight 3): 'abortion', 'gun', 'immigration', 'climate', 'tax', 'healthcare', 'obamacare', 'socialism', 'vaccine', 'blm', 'black lives matter', 'defund', 'wall', 'border'
   \end{itemize}

\item \textbf{Post Scoring:} For each post:
   \begin{itemize}
   \item Count occurrences of each political term in the post text
   \item Multiply each term's count by its assigned weight
   \item Sum these weighted counts to calculate the post's political signal score
   \item Add a small random factor (0-0.01) to break ties between posts with identical scores
   \item Apply a +5 point boost to posts from explicitly political subforums
   \end{itemize}

\item \textbf{Post Selection:} After scoring all posts:
   \begin{itemize}
   \item Sort posts by their political signal scores in descending order
   \item Select 60\% of the required posts from those with highest scores
   \item Select the remaining 40\% to ensure topic diversity, prioritizing posts with different term distributions
   \end{itemize}
\end{enumerate}

This algorithm effectively identifies posts with strong political indicators while maintaining sufficient topical diversity in the selected content for user profile generation.

\section{Additional Figures} \label{appendix:additional-figures}
This appendix contains larger versions of the figures presented in the main text, allowing for more detailed examination.

\clearpage
\begin{figure*}[t]
    \centering
    \includegraphics[width=0.9\textwidth]{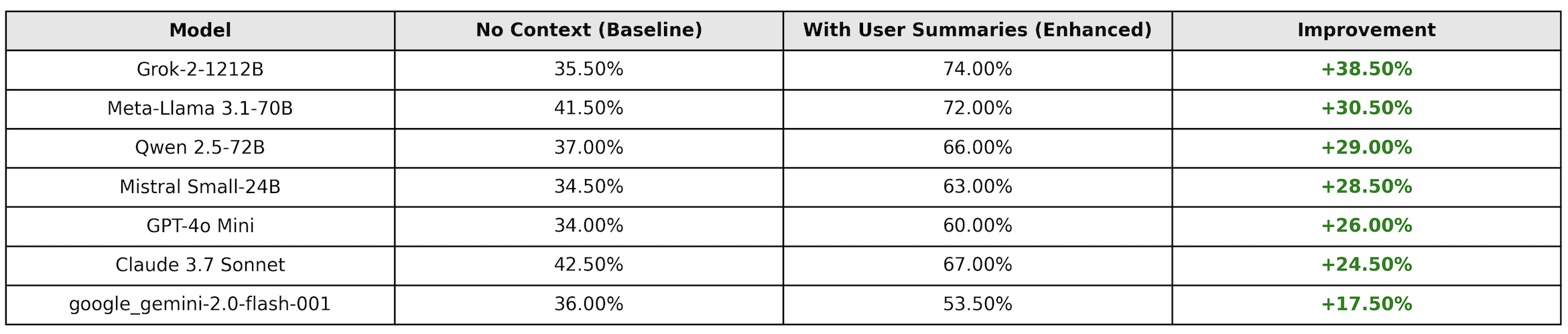}
    \caption*{\large Figure 5: Larger version of Figure 2: Classification accuracy comparison with and without user profile summaries.}
\end{figure*}

\vspace{2cm}

\begin{figure*}[t]
    \centering
    \includegraphics[width=0.9\textwidth]{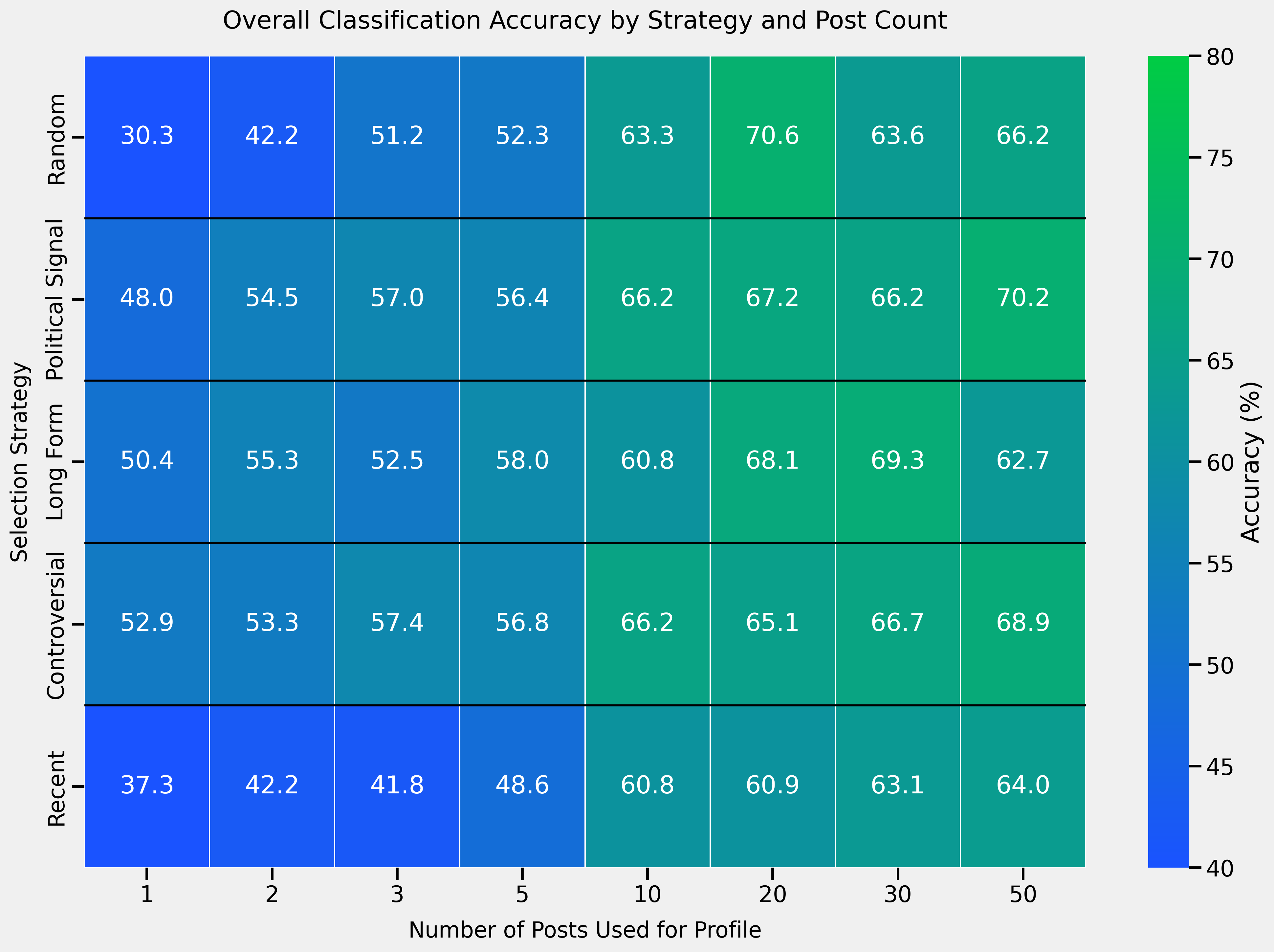}
    \caption*{\large Figure 6: Larger version of Figure 3: Accuracy by post selection strategy and number of posts used for user profiles.}
\end{figure*}

\clearpage

\begin{figure*}[t]
    \centering
    \includegraphics[width=0.9\textwidth]{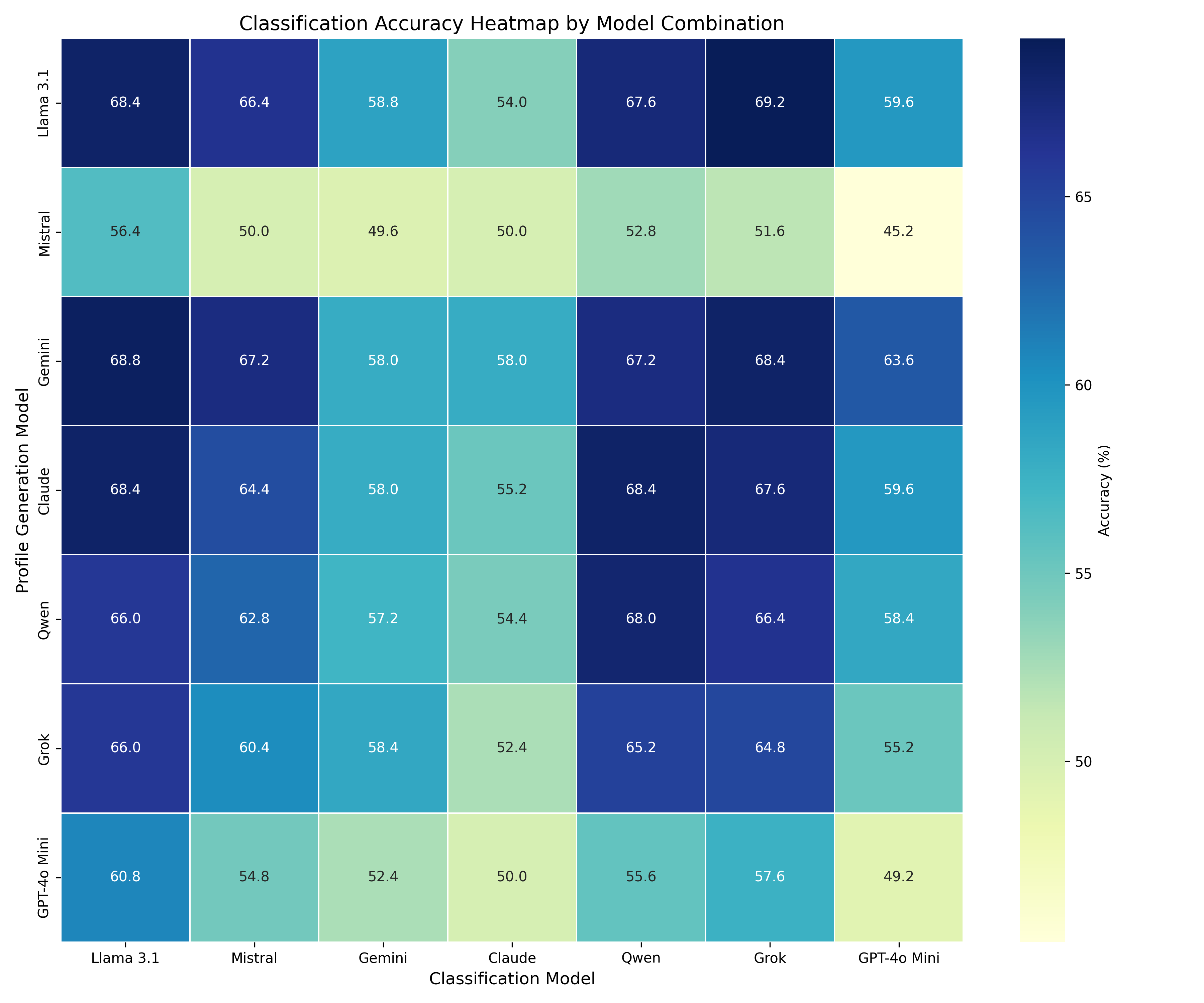}
    \caption*{\large Figure 7: Larger version of Figure 4: Classification accuracy heatmap by model combination.}
\end{figure*}

\end{document}